\documentclass[12pt]{amsart}
\usepackage[margin = 1in]{geometry}
\usepackage{subfig, graphicx, multirow, multicol}
\usepackage{amsmath, amssymb, bm, booktabs}
\usepackage{hyperref}
\usepackage{algorithm}
\usepackage{algpseudocode}
\allowdisplaybreaks

\begin{document}
\title[BertNSP-finance and finbert-lc]{Financial Sentiment Analysis: Leveraging Actual and Synthetic Data for Supervised Fine-tuning}
\author{Abraham Atsiwo}
\address{Department of Mathematics \& Statistics; University of Nevada, Reno}
\email{aatsiwo@unr.edu}

\begin{abstract}
	The Efficient Market Hypothesis (EMH) highlights the essence of financial news in stock price movement. Financial news comes in the form of corporate announcements, news titles, and other forms of digital text. The generation of insights from financial news can be done with sentiment analysis. General-purpose language models are too general for sentiment analysis in finance.  Curated labeled data for fine-tuning general-purpose language models are scare, and existing fine-tuned models for sentiment analysis in finance do not capture the maximum context width. We hypothesize that using actual and synthetic data can improve performance. We introduce \textit{BertNSP-finance} to concatenate shorter financial sentences into longer financial sentences, and \textit{finbert-lc} to determine sentiment from digital text. The results show improved performance on the accuracy and the f1 score for the financial phrasebank data with $50\%$ and $100\%$ agreement levels. 
\end{abstract}

\maketitle

\newtheorem{theorem}{Theorem}
\newtheorem{lemma}{Lemma}
\theoremstyle{definition}
\newtheorem{asmp}{Assumption}
\newtheorem{defn}{Definition}
\newtheorem{example}{Example}
\thispagestyle{empty}

\section{Introduction}
The recent development in natural language processing (\cite{vaswani2017attention}, \cite{devlin2019bertpretrainingdeepbidirectional}, \cite{radford2018improving}) has led to several use cases of textual data: text classification, sentiment analysis, named entity classification, machine translation, text summarization, question answering and text generation. One particular application in finance is sentiment analysis. The efficient market hypothesis highlights the importance of past trading information in stock price movement. This information includes financial news, corporate announcements (layoffs, corporate reports), and other sources of textual data.  Sentiment analysis helps analysts gain insight from digital text. 

Generally speaking, sentiment analysis is a natural language processing technique that defines the emotional tone or sentiment flowing through a particular text. Sentiment analysis classifies digital text into one of three categories: positive, negative, and neutral. 
General-purpose large language models are trained on a vast amount of data from the Web, making it difficult to capture the language used in the financial world. For example, equity used in the general context represents fairness or justice, while equity used in the financial context represents ownership interest in a company. 

Large Language Models (LLMs) can be adopted for financial use cases through in-context learning, prompt engineering, or fine-tuning. Fine-tuning trains the general-purpose large language model on labeled financial data for sentiment analysis. Finbert \cite{araci2019finbert}, Finllama \cite{touvron2023llama} are examples of fine-tuned BERT and LLaMA models, respectively, for sentiment analysis in the financial world. 

Popular data for fine-tuning LLMs for sentiment analysis in finance are the "financial phrasebank" data created by \cite{malo2014good} to bridge the gap between the use cases for LLM in finance. This dataset has a maximum token length of less than 100. BERT and llaMA have a maximum token length of size 512 and 1024 respectively. Fine-tuned models using this dataset do not take advantage of the maximum context window associated with these LLMs, making it difficult to classify sentiment for long financial text. 

We consider a two-step approach to address the issue pertaining to a fine-tuned finance-specific LLM that does not leverage the maximum context width:

\begin{itemize}
	\item[a.] Does augmenting the training data with synthetic data generated from another LLM improve metrics for sentiment analysis?
	\item[b.] Does generating a longer sentences by concatenating the training data sequentially improve performance on Finbert or other fine-tuned language models?  
\end{itemize}

The proposed finbert maximum context (finbert-lc) provides a solution by fine-tuning BERT on the original training data and synthetic data, which captures the maximum context for sentiment analysis in finance. The main contributions are
\begin{itemize}
	\item[a.] A well curated synthetic data for financial sentiment analysis, which captures BERT maximum context window. 
	\item[b.] The proposed models are BertNSP-finance and finbert-lc. We also show that a similar level of performance can be achieved by fine-tuning a smaller number of parameters. 
\end{itemize}

The rest of the article is organized as follows: we present related works in Section 2; model formulation and implementation details in Section 3; main results in Section 4 and the conclusion in Section 5. 
\section{Related Works}

\subsection*{Sentiment Analysis in Finance}

Sentiment analysis, or opinion mining, analyzes digital text to identify and categorize opinions computationally. According to the Efficient Market Hypothesis (EMH) \cite{fama1995random}, both private and public information is reflected in the price of assets. In this case, information refers to financial statements, news reports, analyst recommendations, and private information. The EMH highlights the need to analyze sentiments from digital text. 
Sentiment analysis has three categories: lexicon-based or dictionary-based approach, machine learning approach, and deep learning approach. 

The dictionary-based approach relies on domain-specific information grouped as a dictionary. Given a new sentence, positive and negative words are retrieved from the dictionary and the predicted sentiment class is positive if the number of positive words exceed the number of negative words. The predicted sentiment class is negative otherwise. Studies that employ the lexicon-based approach are discussed in \cite{kanayama2006fully}, \cite{ding2008holistic}, \cite{taboada2011lexicon} and review papers \cite{bonta2019comprehensive}, \cite{sadia2018overview}. 

In \cite{ding2008holistic}, the authors apply the lexicon-based approach to classify the sentiment of customer reviews (positive, negative, or neutral). Their approach is able to handle opinion words that are context-dependent, which poses challenges for some algorithms. H. Kanayama and T. Nasukawa \cite{kanayama2006fully} propose an unsupervised dictionary building technique for the detection of polar clauses. Their approach predicts a negative or a positive class in a specific domain. However, the lexicon-based approach assumes that words in a document are independent. 

Machine learning approaches to sentiment analysis follow the two-step process: 1) numerical representation of text documents and 2) prediction of the sentiment class into positive, negative or neutral. Numerical feature representation can be done using statistical techniques such as the count vectorizer (CV), the term frequency-inverse document frequency (TF-IDF), and word embeddings. CV collapses a sentence into an n-dimensional vector, where $n$ is the vocabulary size. The $i^{th}$ entry in the vector is the token count for the $i^{th}$ word. This approach suppresses important words and gives emphasis to words that are less important, such as prepositions and conjunctions. The TF-IDF approach overcomes the word-vanishing problem of the CV approach. TF-IDF is a product of the TF and the IDF metrics: TF is the number of occurrences of a word in a sentence or document, and IDF penalizes the word count if it appears in more sentences within a text document. The CV and TF-IDF algorithms do not capture the semantics dynamics associated with words, but word embedding does. Words that appear in the same context will not have the same meaning. 

Word embedding is the numerical representation of a token in a document in which words that are similar in meaning and used in the same context are grouped together. 
Word embedding overcomes the lost in semantic meaning that are synonymous with the CV and the TF-IDF algorithms. Examples of word encoders are Word2Vec \cite{mikolov2013distributed}, Glove \cite{pennington2014glove}, FastText \cite{bojanowski2017enriching}, and ELMo \cite{sarzynska2021detecting}. Other encoders are discussed in the review paper \cite{mishev2020evaluation}. Machine learning algorithms used with CV, TF-IDF and embeddings for classifying human sentiments are Support Vector Machines (SVM) \cite{martineau2009delta}, Naives Bayes (NB), Maximum Entropy (ME), Stochastic Gradient Descent (SGD) \cite{tripathy2016classification}. 

Deep learning techniques for sentiment analysis follow the same two-step process. The machine learning model in step 2 is replaced by a deep learning model. One of the earlier papers that used deep learning for financial analysis is "Decision support from financial disclosures with deep neural networks and transfer learning" \cite{kraus2017decision}. They predict stock market movements with long-short-term memory (LSTM) in company announcements and conclude that the LSTM method is better than traditional machine learning approaches. The paper also experiments with transfer learning, where the LSTM network is pre-train on a different corpus with 139.1 million words. \cite{sohangir2018big} applies several neural network architectures, such as LSTM, doc2vec, and the convolutional neural network, to predict the sentiment of a large group of financial experts towards a stock. Experiments show that the convolutional neural network is the best architecture for financial sentiment analysis with the StockTwits dataset. In \cite{maia2018finsslx}, the authors introduce FinSSLx, a sentiment-based prediction model for the financial domain. The authors employ a sentiment simplification step, where complex sentences are broken down into shorter sentences that are then classified according to the polarity. 

Machine learning approaches do not capture the sequential nature associated with language. Deep learning techniques are data-hungry. Most financial companies do not release their financial data set to the public. Open source data for financial use cases are in short supply, making it difficult to apply deep learning approaches that show promising results. 

\subsection*{Pre-Trained and Finetuned Models for Text Classification}
The transformer architecture introduced by Google research \cite{vaswani2017attention} in the paper "Attention is all you need" has revolutionized language modeling. Language modeling is the art of predicting the next word or surrounding words given a sequence of text. 
The transformer architecture, unlike CNN and the LSTM architecture, is parallelizable. It uses an attention mechanism to track the order of words, eradicating the major flaw of traditional machine learning approaches. Examples of language models that use the transformer architecture are BERT \cite{devlin2019pre}, GPT \cite{radford2018improving}, and LLaMa \cite{touvron2023llama}. These models are trained on diverse corpus and do not generalize well in the financial domain. There are not enough data to train a language model from scratch for use in the financial domain.

Bloombergpgt \cite{wu2023bloomberggpt} is the first finance-specific large language model in the literature. This LLM has 50 billion parameters and was trained on a 365 billion internal Bloomberg token data source. The model was trained for 139,200 steps, which took approximately 53 days. It is used for sentiment analysis, name-entity recognition, named-entity disambiguation, and knowledge assessments. The cost to train such a model, despite its performance on various tasks, is enormous (1.3M GPU hours on 40GB A100 GPUs). 

Language models can be adopted for task-specific use cases by instruction fine-tuning. Embeddings from Language Models (ELMo)\cite{sarzynska2021detecting} is the first language model to implement finetuning. In their implementation, the authors used a CNN to generate word embedding vectors and then two layers of bidirectional LSTM for human clinical ratings. The CNN and the LSTM are pre-trained on a large corpus on text and adopted for human clinical ratings. 

FinBERT \cite{araci2019finbert} is a fine-tuned BERT model for sentiment analysis in the financial domain. The financial phrasebank \cite{malo2014good} dataset was used for fine-tuning and achieved state-of-the-art performance on various metrics compared to the BERT base model, machine learning, and deep learning approaches. Other models fine-tuned for financial use cases are FinGPT, \cite{liu2023fingpt}, Instruct-fingpt \cite{zhang2023instruct}.

Despite the remarkable success of FinBERT for sentiment analysis, we want to take a different path. In \cite{tang2023does}, the authors show the effectiveness of LLM synthetic data generation for the extraction of clinical text. AugGPT \cite{dai2023auggpt} augments the data to capture the invariance of the data and generate more samples. Synthetic data generation helps ease the privacy concerns associated with certain fine-tuning applications. It enables us to generate data of different lengths to capture various dynamics in real-world applications. We would like to pursue this path: augment the training data with synthetic data and then fine-tune BERT for sentiment analysis in the financial domain. 

In conclusion, this comprehensive examination of the existing literature lays the foundation for augmenting the training data with data generated by another LLM to improve text classification metrics in the financial domain.
\section{Experimental Setup and Analysis}\label{finetuning::experimental_setup}
We present the implementation details and configuration in this section. We discuss the datasets used in the analysis, baseline models, evaluation metrics, implementation details, and main results. 
\subsection{Data Sources}
The BERT model for next seqence prediction was further pre-trained with data from Bloombery compiled by \cite{BloombergReutersDataset2015}. The data span seven years, from October 2006 to November 2013. It is not possible to regenerate or update the data because Bloombery shut down their public API years ago. For a given pair of sentences (sentenceA, sentenceB). The sentences were generated by splitting a paragraph or a long sentence into two parts. The positive class "isNext" was taken as the actual next sentence, and the negative class "notNext" was randomly sampled from other sentence pairs, resulting in approximately eight million sentence pairs: four million positive examples and four million negative examples. Only 20,000 examples were used in further pre-training the BERT model for next sequence prediction.

Most data are close-sourced by financial institutions, making it difficult to access them for open-source implementation of machine learning and deep learning models. \cite{malo2014good} in the paper "Detecting Semantic Orientations in Economic Texts" bridges this gap, at least in sentiment analysis, by curating human-annotated data available to researchers. Financial phrasebank, in the literature, points to these curated data. The data distribution is summarized in Table \ref{finetuning::experimental_setup::table::financial-phrasebank}.

\begin{table}[!h]
\centering
\begin{tabular}{@{}lllll@{}}
%\toprule
                                & \% Negative & \% Neutral & \% Positive & Count \\ \midrule
Sentences with $100\%$ agreement & 13.4        & 61.4       & 25.2        & 2259  \\
Sentences with $75\%$ agreement  & 12.2        & 62.1       & 25.7        & 3448  \\
Sentences with $66\%$ agreement  & 12.2        & 60.1       & 27.7        & 4211  \\
Sentences with $50\%$ agreement  & 12.5        & 59.4       & 28.2        & 4840  \\ 
%\bottomrule
\end{tabular}
\caption{Distribution of labels in financial phrasebank for four subsets formed based on the strength of majority agreement. Each sentence has five to eight overlapping annotations, which have been used to determine the degree of agreement \cite{malo2014good}.}
\label{finetuning::experimental_setup::table::financial-phrasebank}
\end{table}

\begin{figure}[!h]
    \centering
    \includegraphics[width=1\linewidth, height=0.5\linewidth]{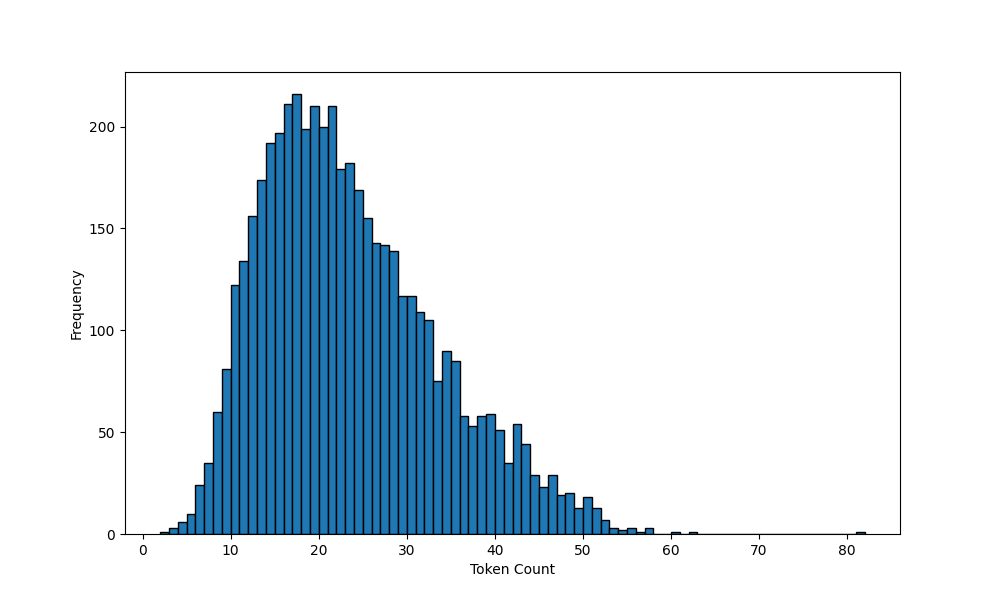}
    \caption{Token distribution of the financial phrasebank dataset.}
    \label{finetuning::experimental_setup::Figure::financial-phrasebank-short}
\end{figure}

\begin{figure}[!h]
    \centering
    \includegraphics[width=1\linewidth, height=0.5\linewidth]{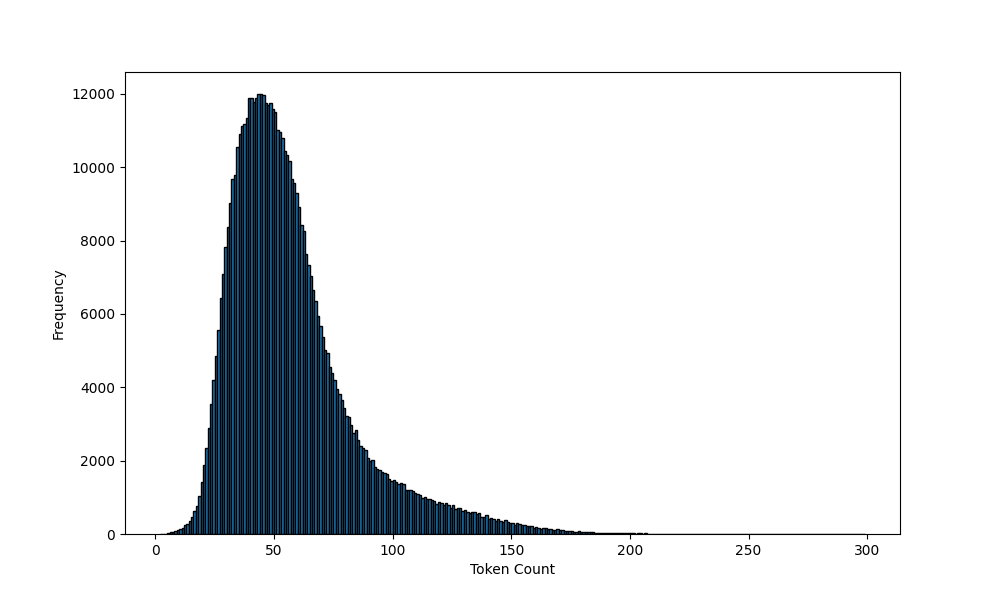}
    \caption{Token distribution of the concatenated financial phrasebank dataset.}
    \label{finetuning::experimental_setup::Figure::financial-phrasebank-short-long}
\end{figure}

Long sentences (financial phrasebank concatenate) were generated with two approaches: random concatenation and sequential concatenation. Sequential concatenation was implemented with Algorithm \ref{finetuning::experimental_setup::algorithm::long-sentence}. Random concatenation involves selecting positive (negative or neutral) sentences and concatenating them into a longer sentence. The distribution of financial phrasebank and financial phrasebank concatenate is presented in Figures \ref{finetuning::experimental_setup::Figure::financial-phrasebank-short} and \ref{finetuning::experimental_setup::Figure::financial-phrasebank-short-long}, respectively.

It can be seen from the histograms that the maximum token count is 82 and the minimum token count is 2 for the original dataset and the maximum is 298 and the minimum is 2 for the concatenated and GPT-3 generated datasets. The BERT model has a maximum token count of 512. 

Training examples were generated from the fine-tuned GPT-3 model. This supplements the financial phrasebank, financial phrasebank concatenate training data. 

\begin{algorithm}
\caption{predict\_multiple\_nsp}
\label{finetuning::experimental_setup::algorithm::long-sentence}
\begin{algorithmic}[1]
\Procedure{predict\_multiple\_nsp}{sentences: List[str], concatenate: bool}
    \State valid $\gets$ \textbf{True}
    \For{$i \gets 0$ to $|\texttt{sentences}| - 1$}
        \If{concatenate}
            \State sentenceA $\gets$ \texttt{concatenate\_sentence}(sentences[0 : i+1])
            \State sentenceB $\gets$ sentences[i + 1]
        \Else
            \State sentenceA $\gets$ sentences[i]
            \State sentenceB $\gets$ sentences[i + 1]
        \EndIf
        \If{\Call{predict\_nsp}{sentenceA, sentenceB} $\leq$ 0.5}
            \State valid $\gets$ \textbf{False}
            \State \textbf{break}
        \EndIf
    \EndFor
    \State \Return valid
\EndProcedure
\end{algorithmic}
\end{algorithm}

\subsection{Baseline Methods and Evaluation Metrics}
We compare the proposed model with FINBERT (Financial Sentiment Analysis with Pre-trained Language Models) and Long-Short-Term Memory (LSTM). A bidirectional LSTM with four embedding vectors: ELMO, BERT, GLOVE, and FASTTEXT embeddings. 
Different configurations are tried with these embeddings, and the configuration with the best result is reported for each model. The bidirectional LSTM has a hidden size of 128 (forward) and 128 (backward) neurons, max pooling, layer normalization, and a fully connected layer with three neurons. The last hidden layer has 256 (128 forward + 128 backward) neurons. Pooling reduces the dimensionality of the output from the bidirectional LSTM and layer normalization stabilizes and speeds up the training process. 

FINBERT weights were downloaded from the Github repository referenced in the original paper \cite{araci2019finbert}. The weights can also be downloaded through the hugging face (an open-souce community focused on advancing technology for the advancement in natural language processing) platform.

We used test loss, accuracy, and f1 score for comparison. Accuracy is the ratio of the correctly predicted response to the number of instances, and the macro f1 score is the harmonic mean of macro precision and macro recall. The f1 score is used when the data are imbalanced among classes, which is the case in this application. 

% \subsection*{Evaluation Metrics}
\subsection{Implementation Details}
BERT was fine-tuned with a learning rate of $2e^{-5},$ a maximum token length of 512, a mini-batch size of 8, a weight decay of $0.01$, a dropout probability of 0.2, an exponential decay rate for the first moment estimates $(\beta_1)$ of $0.9$, an exponential decay rate for the second moment estimates $(\beta_2)$ of $0.999.$ An Macbook pro (Apple Silicon with 16GB ram and 10 cores GPU) is used to train, validate and test all the LSTM models. A “4x A6000 (48 GB)” configuration on Lambda Labs (a cloud platform for training and fine-tuning large language models) is used to fine-tune all BERT models. This graphics card has four NVIDIA RTX A6000 GPUs, each with 48 GB of memory, which provides exceptional computational power and memory capacity to fine-tune AI, deep learning, and large language models. 

The embedding, the twelve encoder blocks and the output layer are frozen one at a time. Freezing reduces the number of parameters to be trained, which subsequently reduces the training time. We seek the layer with the greatest impact on accuracy and f1 score.  See Chapter 2 in \cite{atsiwo2024instruction} for the details on model training. 

\section{Experimental Results}
In this section, we present results from different experiments. First, we compare the performance of the pre-trained BERT (BertNSP-finance) on Bloombery data to Vanilla BERT (no further pre-training). Next, we compare finbert-lc with FINBERT, Bidirectional LSTM with the four embeddings (ELMO, GLOVE, FASTTEXT and BERT). Lastly, we discuss where finbert-lc and BertNSP-finance fail by examining the confusion matrix. We evaluated models with accuracy and f1 score for the financial phrasebank dataset with different agreement levels. 

The model weights for BertNSP-finance and finbert-lc can be downloaded from \\ \textit{https://huggingface.co/ab30atsiwo/nsp-finetuned-bloombery} and \\ \textit{https://huggingface.co/ab30atsiwo/finbert-gpt-50agree}, respectively. The code can be downloaded from the github repository \textit{https://github.com/abraham-atsiwo/filbert-lc}.

\subsection{Effects of Further Pretraining BERT for NSP in the financial Domain}
BertNSP-finance is trained on bloombery data to predict the next sentence in the financial domain. Vanilla BERT is trained on a general corpus. 

\begin{table}[h!]
\centering
\begin{tabular}{@{}lcccc@{}} \toprule
Model      & \multicolumn{1}{l}{\# parameters} & \multicolumn{1}{l}{loss} & \multicolumn{1}{l}{accuracy} & \multicolumn{1}{l}{f1 macro} \\ \midrule
Vanilla BERT Small & 110M                             & 0.61                   & 0.77                     & 0.76                      \\
Vanilla BERT Large & 340M                              & 0.67                & 0.62                     & 0.57                       \\
BertNSP-finance  & 110M                              & \textbf{0.24}                  & \textbf{0.91}                       & \textbf{0.91}                       \\ \bottomrule
\end{tabular}
\caption{Loss, f1-score and accuracy score for finetuned BERT for next sequence prediction for a test data of size 5000. Loss (small is better), accuracy and f1-score (large is better). The model with \textbf{Bold face} indicates the best results in the corresponding metric.}
\label{finetuning::experimental_setup::table::nsp_metrics}
\end{table}

For all metrics considered as shown in Table \ref{finetuning::experimental_setup::table::nsp_metrics}, BertNSP-finance outperformed Vanilla BERT Small and Vanilla BERT Large. Vanilla BERT Large is the worst for all metrics. This experiment shows that using a model with a high number of parameters does not always give the best results. 

\begin{figure}[h!]
    \centering
    \includegraphics[width=1.1\linewidth, height=0.7\linewidth]{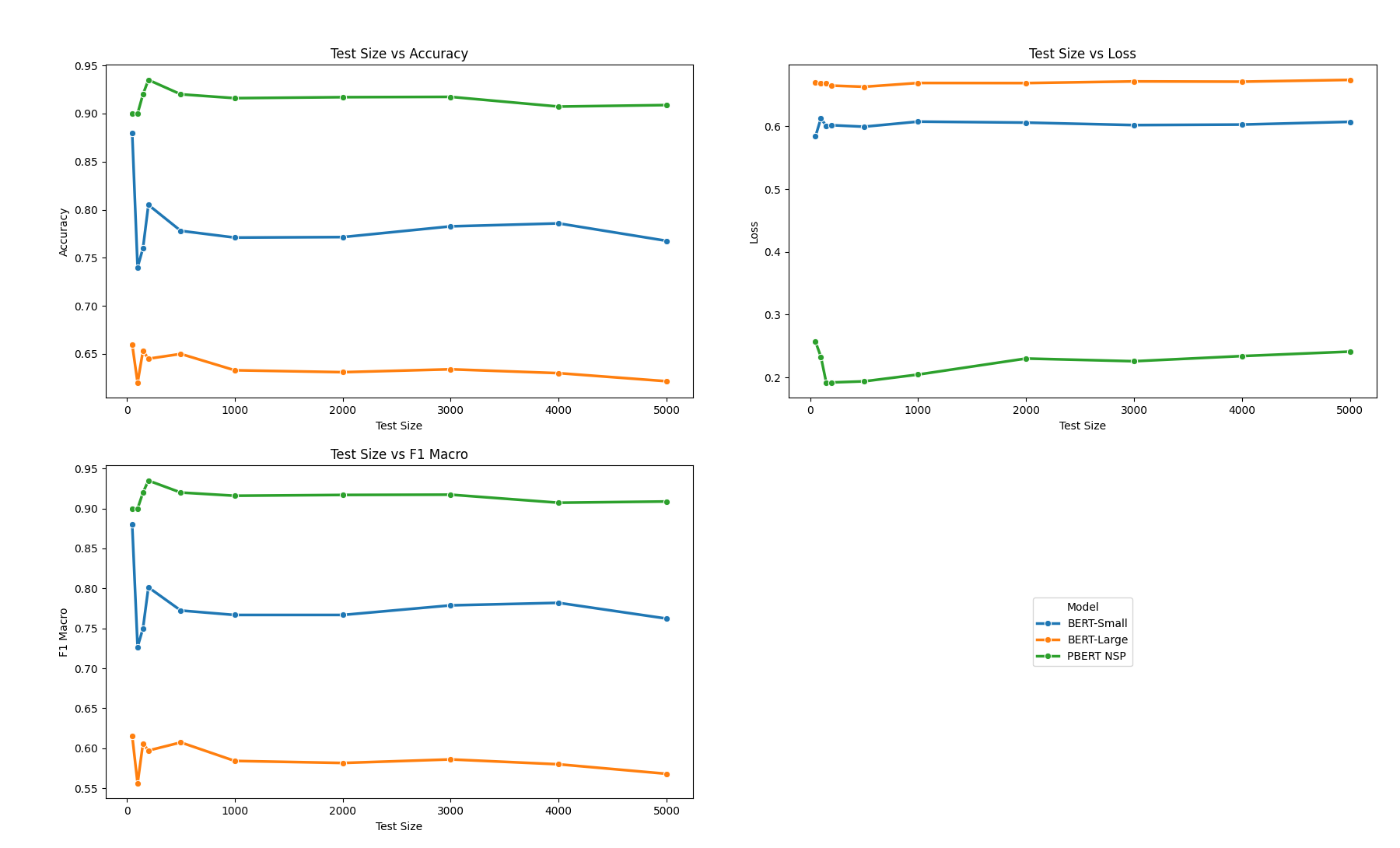}
    \caption{Plot of test size vs. accuracy, loss and vs. f1 macro grouped by model type (Vanilla BERT Small, Vanilla BERT Large and BertNSP-finance (PBERT NSP)).}
    \label{finetuning::experimental_setup::figure::nsp_metrics}
\end{figure}

Vanilla BERT Large has a precision of $57\%$ for the positive class (isNext) and a recall of $27\%$ for the negative class (notNext). A precision of $57\%$ means that out of all the instances predicted as "isNext", "sentenceB" follows "sentenceA" $57\%$ of the time. $57\%$ of the "isNext" predictions made by the model are correct. Recall of $37\%$ suggests that the model fails to identify negative instances (notNext) $63\%$ of the time, which means that there are a significant number of false positives. F1 macro is calculated with precision and recall, explaining why Vanilla BERT large has an f1 macro score of $57\%.$ On the other hand, BertNSP-finance has a precision of $91\%$ for the positive class and a recall of $91\%$ for the negative class. It has the same percentage for the positive class (recall) and the negative class (precision), explaining the reason for an f1 macro score of $91\%$ for the model further pre-trained. If data are not available for further pretraining, then Vanilla BERT Small is a better option than Vanilla BERT Large for sentence-pair prediction.

Figure \ref{finetuning::experimental_setup::figure::nsp_metrics} is a plot of accuracy, loss, f1 macro for each model as the length of the test data is varied. 
It can be seen from Figure \ref{finetuning::experimental_setup::figure::nsp_metrics} that BertNSP-finance has the lowest loss, the highest accuracy, and f1 micro for all the test sizes considered. The loss for BERT (Small and Large) is somehow constant for different test sizes; It is constant for BertNSP-finance after a test data of size 3000.

\subsection{finbert-lc, FINBERT, and LSTM for Text Classification}
We compare the performance of finbert-lc, FINBERT, and LSTM for text classification in the financial domain. In some cases results are reported for the financial phrasebank data with all agreement levels. The test loss, accuracy, and f1 score are reported for all models. The metrics are summarized in Table \ref{finetuning::experimental_setup::table::metrics_lstm_finbert-50agree}.

\begin{table}[h!]
\centering
\begin{tabular}{@{}lllll@{}}
\toprule
\multicolumn{2}{l}{Model}                                             & \multicolumn{3}{l}{Metrics} \\ \midrule
Type                                                     & Name       & Loss  & Accuracy & F1 Score \\ \midrule
\multicolumn{1}{r}{\multirow{4}{*}{LSTM With Embedding}} & BERT       & 0.42  & 0.83     & 0.79     \\
\multicolumn{1}{r}{}                                     & GLOVE      & 0.69  & 0.81     & 0.77     \\
\multicolumn{1}{r}{}                                     & FASTTEXT   & 0.46  & 0.82     & 0.78     \\
\multicolumn{1}{r}{}                                     & ELMO       & 0.42  & 0.82     & 0.82     \\ \midrule
\multirow{2}{*}{BERT}                                    & FINBERT    & \textbf{0.37}  & 0.86     & 0.84     \\
                                                         & \textbf{finbert-lc} & 0.43  & \textbf{0.89}     & \textbf{0.88}     \\ \midrule
\end{tabular}
\caption{Loss, accuracy and f1 score for LSTM with four different embeddings and fine-tuned BERT models for the financial phrasebank data with $50\%$ agreement and GPT-4 generated data.}
\label{finetuning::experimental_setup::table::metrics_lstm_finbert-50agree}
\end{table}

\begin{table}[h!]
\centering
\begin{tabular}{@{}lllll@{}}
\toprule
\multicolumn{2}{l}{Model}                                             & \multicolumn{3}{l}{Metrics} \\ \midrule
Type                                                     & Name       & Loss  & Accuracy & F1 Score \\ \midrule
\multicolumn{1}{r}{\multirow{4}{*}{LSTM With Embedding}} & BERT       & 0.21  & 0.91     & 0.83     \\
\multicolumn{1}{r}{}                                     & GLOVE      & 0.26  & 0.93     & 0.90    \\
\multicolumn{1}{r}{}                                     & FASTTEXT   & 0.28  & 0.93     & 0.90     \\
\multicolumn{1}{r}{}                                     & ELMO       & 0.19  & 0.92     & 0.93  \\ \midrule
\multirow{2}{*}{BERT}                                    & FINBERT    & \textbf{0.13}  & \textbf{0.97}     & 0.95     \\
                                                         & \textbf{finbert-lc} & \textbf{0.13}  & \textbf{0.97}     & \textbf{0.96}     \\ \midrule
\end{tabular}
\caption{Loss, accuracy and f1 Score for LSTM with different four different embeddings and finetuned BERT models for the financial phrasebank data with $100\%$ agreement and GPT-4 generated data}
\label{finetuning::experimental_setup::table::metrics_lstm_finbert-100agree}
\end{table}

With the data with $50\%$ agreement level Table \ref{finetuning::experimental_setup::table::metrics_lstm_finbert-50agree}, FINBERT has the lowest loss and finbert-lc has the highest accuracy and f1 score. The data are imbalanced, and thus loss and f1-score provide a better measure of performance than accuracy. The LSTM implementation with pre-trained BERT, GLOVE, FASTTEXT and ELMO embeddings have a higher f1 score than anticipated. To my knowledge, this is one of the best implementations of LSTM with pre-trained embeddings for sentiment analysis in the financial domain in terms of reported metrics. The loss is on the higher side for our fine-tuned model, finbert-lc, despite a reported f1-score of $89\%.$

With the data with $100\%$ agreement level Table \ref{finetuning::experimental_setup::table::metrics_lstm_finbert-100agree} the same loss is reported for FINBERT and finbert-lc, followed by LSTM with ELMO embeddings, and LSTM with FASTTEXT embeddings has the largest loss. FINBERT and finbert-lc also have the same accuracy for these data, and LSTM with BERT embeddings has the smallest. There is a $1\%$ improvement in the f1 score by augmenting the financial phrasebank data with the GPT4 generated data. In general, enhancing the data with GPT4 generated improved all metrics other than loss for data with $50\%$ agreement level and achieved the same or better values in terms of metrics for data with $100\%$ agreement level.

\subsection{Measuring Performance by Freezing Layers}
\begin{table}[h!]
\centering
\begin{tabular}{lcccc}
\toprule
Freezed Layer & ~$\#$ Trainable Parameters & Loss & Accuracy & F1 Score \\
\midrule
Embedding Layer & 86M & 0.43 & 0.88 & 0.88 \\
Layer 1 & 79M & 0.45 & 0.88 & 0.88 \\
Layer 2 & 71M & 0.43 & 0.87 & 0.87 \\
Layer 3 & 64M & 0.42 & 0.88 & 0.87 \\
Layer 4 & 57M & 0.42 & 0.87 & 0.87 \\
Layer 5 & 50M & 0.39 & 0.88 & 0.88 \\
Layer 6 & 43M & 0.39 & 0.87 & 0.87 \\
Layer 7 & 36M & 0.36 & 0.87 & 0.86 \\
Layer 8 & 29M & 0.36 & 0.87 & 0.86 \\
Layer 9 & 22M & 0.36 & 0.87 & 0.86 \\
Layer 10 & 15M & 0.33 & 0.88 & 0.86 \\
Layer 11 & 8M & 0.38 & 0.83 & 0.80 \\
Layer 12 & 0.5M & 0.71 & 0.70 & 0.52 \\
\bottomrule
\end{tabular}
\caption{Number of trainable parameters, loss, accuracy, and f1-Score for each layer when freezing all layers up to and including the specified layer for sentences with $50\%$ agreement.}
\label{finetuning::experimental_setup::table::freezing_layer-50agree}
\end{table}

Next, we explore the impact of freezing different parts (encoder blocks and embedding layer) of the model architecture Table \ref{finetuning::experimental_setup::table::freezing_layer-50agree} on the metrics reported. Freezing reduces the number of parameters to be fine-tuned, subsequently decreasing the time complexity of fine-tuning. Freezing the embedding layer implies that we train only parameters in layers 1 through 12 and in the classification layer. Freezing layer 5 implies that the embedding layer and layers 1 through 5 are frozen. We train the parameters in layers 6 through 12 and in the classification layer. Freezing layer 12 is equivalent to fine-tuning parameters in the classification layer. We tend to train fewer parameters as we freeze more layers. Freezing layers 1 through 10 does not affect the reported metrics as shown in Table \ref{finetuning::experimental_setup::table::freezing_layer-50agree}. There is a $5\%$ decrease in accuracy and a $6\%$ decrease in the f1 score of layer 10 by freezing layer 11. Fine-tuning the output layer gives an f1 score of $52\%$, an accuracy of $70\%$ and a loss of $71\%$. The trade-off here is the number of parameters to be trained. 

State-of-the-art performance can be achieved by training 15M parameters instead of 110M parameters in the original BERT base implementation. Achieving state-of-the-art performance by fine-tuning 15M parameters is quite remarkable, especially for a large language model.

% \subsubsection{LSTM, FINBERT, and finbert-lc on Test Data with Different Length}

\subsection{Confusion Matrix and where the Model Fails}
There are failures that need to be highlighted despite the state-of-the-art performance achieved by finbert-lc and BertNSP-finance in the reported metrics. The confusion matrices are reported in Tables \ref{finetuning::experimental_setup::table::confusion_matrix_50agree} and \ref{finetuning::experimental_setup::table::confusion_matrix_allagree}.

\begin{table}[h!]
\centering
\begin{tabular}{cc|c|c|c|}
\cline{3-5}
 & & \multicolumn{3}{c|}{\textbf{Predicted}} \\
\cline{3-5}
 & & Negative & Neutral & Positive \\
\hline
\multicolumn{1}{|c|}{\multirow{3}{*}{\textbf{Actual}}} & Negative & 53 & 5 & 2 \\
\multicolumn{1}{|c|}{} & Neutral & 7 & 263 & 18 \\
\multicolumn{1}{|c|}{} & Positive & 0 & 23 & 114 \\
\hline
\end{tabular}
\caption{finbert-lc: confusion matrix for financial phrasebank data with 50\% Agreement}
\label{finetuning::experimental_setup::table::confusion_matrix_50agree}
\end{table}
For the data with $50\%$ agreement level, no positive instance was incorrectly classified as negative. $23$ positive examples are classified as neutral. 7 negatives, 25 neutral, and 23 positive sentences are incorrectly classified. 

\begin{table}[h!]
\centering
\begin{tabular}{cc|c|c|c|}
\cline{3-5}
 & & \multicolumn{3}{c|}{\textbf{Predicted}} \\
\cline{3-5}
 & & Negative & Neutral & Positive \\
\hline
\multicolumn{1}{|c|}{\multirow{3}{*}{\textbf{Actual}}} & Negative & 27 & 1 & 2 \\
\multicolumn{1}{|c|}{} & Neutral & 0 & 138 & 1 \\
\multicolumn{1}{|c|}{} & Positive & 1 & 1 & 55 \\
\hline
\end{tabular}
\caption{finbert-lc: Confusion matrix for financial phrasebank data with 100\% Agreement}
\label{finetuning::experimental_setup::table::confusion_matrix_allagree}
\end{table}

The data with $100\%$ agreement have $3, 1$ and $2$ negative, neutral, and postive misclassified sentences, respectively. No neutral instance was classified as negative, 1 positive instance was classified as negative, and 2 negative instances are classified as positive. Details of two misclassified neutral sentences are discussed in the following.

Consider the following two neutral sentences, which are predicted as negative and positive, respectively, in Example \ref{finetuning::experimental_setup::example::neutral_misclassified}.
\begin{example}[Neutral predicted as negative / positive] 
\label{finetuning::experimental_setup::example::neutral_misclassified}
\end{example}
% \vspace{-2em}

\begin{itemize}
    \item[1.] \textit{Rosen was cautious about being too optimistic inregard to the second half of the year.} (\textbf{Neutral} Predicted as \textbf{Negative})
    \item[2.]  \textit{The acquisition is part of Suomen Helasto 's strategy to expand the LukkoExpert Security chain , Suomen Helasto CEO Kimmo Uusimaki said.} (\textbf{Neutral} Predicted as \textbf{Positive})
\end{itemize}

The occurrence of positive, negative, or neutral words in a sentence is likely to have an impact on the predicted class. Misclassification occurs if the model is unable to identify the correct class using the context in which it occurs. For example, volatile can be used in a neutral and a negative context. \textit{“The stock market is volatile, with frequent fluctuations in prices'} has a neutral connotation, and \textit{ 'Investing in this sector is risky because it is highly volatile and unpredictable'} has a negative connotation.

In Example \ref{finetuning::experimental_setup::example::neutral_misclassified}(1), "cautious" is used in a neutral context, but the model incorrectly classified it a negative, and "expand" used in Example \ref{finetuning::experimental_setup::example::neutral_misclassified} (2) also appears in a neutral context, but the model incorrectly classified it as positive. The financial phrasebank data, when classified by human annotators, had different agreement levels. Despite the misclassification in the confusion matrix, the model correctly predicted most instances correctly ($89\%$ accuracy for sentences with $50\%$ agreement and $97\%$ precision for sentences with $100\%$ agreement).

Table \ref{finetuning::experimental_setup::example::nsp_misclassified} is the confusion matrix of fine-tuned BERT for next sentence prediction in the financial domain (PBERTGPT). 223 instances have sentenceB not "following" sentenceA, but was incorrectly classified. On the other hand, 233 instances have sentenceB "following" sentenceA but were classified otherwise. 

\begin{table}[h]
\centering
\begin{tabular}{cc|c|c|}
\cline{3-4}
 & & \multicolumn{2}{c|}{\textbf{Predicted}} \\
\cline{3-4}
 & & notNext & isNext \\
\hline
\multicolumn{1}{|c|}{\multirow{2}{*}{\textbf{Actual}}} & notNext & 2277 & 223 \\
\multicolumn{1}{|c|}{} & isNext & 233 & 2267 \\
\hline
\end{tabular}
\caption{FBERTNSP: Confusion matrix for fine-tuned BERT for next sentence prediction}
\end{table}
We consider the two examples, where "isNext" is predicted as "notNext" and "notNext" is predicted as "isNext" respectively. 

\begin{example}[Misclassified Next Sentence Prediction] 
\label{finetuning::experimental_setup::example::nsp_misclassified}
\end{example}
\begin{itemize}
        \item[1.] \textbf{notNext} predicted as \textbf{isNext} \\
        \textbf{SentenceA}: \textit{BBM, which comes already installed on BlackBerry phones,now has more than 80 million users, including 20 million onGoogle Inc.’s Android platform and Apple Inc.’s iOS, BlackBerrysaid last month.} \\
\textbf{SentenceB}: \textit{
There would be many implementation challenges in combining the programs, they said.}

\item[2.] \textbf{isNext} predicted as \textbf{notNext} \\
\textbf{SentenceA}: \textit{Without reporting flight plans or identifying themselves,} \\
\textbf{SentenceB}: \textit{
Japan lodged a complaint as the US and South Korea expressed concern about China’s actions.}
\end{itemize}

In Example \ref{finetuning::experimental_setup::example::nsp_misclassified} (1), based on the given context, sentenceB does not logically follow sentenceA, but the model predicted otherwise. SentenceA discusses the usage statistics and cross-platform availability of BBM (BlackBerry Messenger), while sentenceB suddenly shifts to talking about implementation challenges in combining programs without any apparent connection to the first sentence.

In Example \ref{finetuning::experimental_setup::example::nsp_misclassified} (2), the two sentences are related, but the model predicted that they are not related. This can be caused by sentenceA not being complete.

\section{Conclusion and Further Work}
We implemented BertNSP-finance and finbert-lc. BertNSP-finance predicts whether sentenceB follows sentenceB, helping to generate longer financial sentiments. BertNSP-finance predicts 1 if sentenceB follows sentenceA and 0 otherwise. Synthetic data generated from another large language model, longer sentences genenerated with BertNSP-finance, and the financial phrasebank data were used to fine-tune BERT for sentiment analysis in finance, achieving state of the art performance compared with other models in selected metrics.  

Finbert-lc performed better in terms of accuracy and f1 score compared to LSTM with BERT, GLOVE, FASTTEXT, and ELMO embeddings for sentences with levels of agreement $50\%$ and $100\%$.  

BERT was further pre-trained on bloombery data before fine-tuned on labeled data for sentiment analysis in finance. Further pretraining of BERT on bloomery data does not improve performance compared to BERT in its natural state. Sentiment analysis is not used on its own. It is used in stock market analysis as part of the capital asset pricing model framework \cite{elbannan2015capital}, \cite{atsiwo2024capitalassetpricingmodel},  algorithmic trading, customer insights in banking, and other use cases.

%\clearpage
%\bibliographystyle{plain}
%\bibliographystyle{alpha}
%\bibliography{references}

\begin{thebibliography}{10}

\bibitem{araci2019finbert}
Dogu Araci.
\newblock Finbert: Financial sentiment analysis with pre-trained language
  models.
\newblock {\em arXiv preprint arXiv:1908.10063}, 2019.

\bibitem{atsiwo2024instruction}
Abraham Atsiwo.
\newblock {\em Instruction Finetuning Foundation Models, Three-Stage Bubble
  Analysis, and Examining the Size Effect}.
\newblock PhD thesis, University of Nevada, Reno, 2024.

\bibitem{atsiwo2024capitalassetpricingmodel}
Abraham Atsiwo and Andrey Sarantsev.
\newblock Capital asset pricing model with size factor and normalizing by
  volatility index, 2024.

\bibitem{bojanowski2017enriching}
Piotr Bojanowski, Edouard Grave, Armand Joulin, and Tomas Mikolov.
\newblock Enriching word vectors with subword information.
\newblock {\em Transactions of the association for computational linguistics},
  5:135--146, 2017.

\bibitem{bonta2019comprehensive}
Venkateswarlu Bonta, Nandhini Kumaresh, and Naulegari Janardhan.
\newblock A comprehensive study on lexicon based approaches for sentiment
  analysis.
\newblock {\em Asian Journal of Computer Science and Technology}, 8(S2):1--6,
  2019.

\bibitem{dai2023auggpt}
Haixing Dai, Zhengliang Liu, Wenxiong Liao, Xiaoke Huang, Yihan Cao, Zihao Wu,
  Lin Zhao, Shaochen Xu, Wei Liu, Ninghao Liu, et~al.
\newblock Auggpt: Leveraging chatgpt for text data augmentation.
\newblock {\em arXiv preprint arXiv:2302.13007}, 2023.

\bibitem{devlin2019pre}
J~Devlin, MW~Chang, K~Lee, and KB~Toutanova.
\newblock Pre-training of deep bidirectional transformers for language
  understanding in: Proceedings of the 2019 conference of the north american
  chapter of the association for computational linguistics: Human language
  technologies, volume 1 (long and short papers).
\newblock {\em Minneapolis, MN: Association for Computational Linguistics},
  pages 4171--86, 2019.

\bibitem{devlin2019bertpretrainingdeepbidirectional}
Jacob Devlin, Ming-Wei Chang, Kenton Lee, and Kristina Toutanova.
\newblock Bert: Pre-training of deep bidirectional transformers for language
  understanding, 2019.

\bibitem{ding2008holistic}
Xiaowen Ding, Bing Liu, and Philip~S Yu.
\newblock A holistic lexicon-based approach to opinion mining.
\newblock In {\em Proceedings of the 2008 international conference on web
  search and data mining}, pages 231--240, 2008.

\bibitem{elbannan2015capital}
Mona~A Elbannan.
\newblock The capital asset pricing model: an overview of the theory.
\newblock {\em International Journal of Economics and Finance}, 7(1):216--228,
  2015.

\bibitem{fama1995random}
Eugene~F Fama.
\newblock Random walks in stock market prices.
\newblock {\em Financial analysts journal}, 51(1):75--80, 1995.

\bibitem{kanayama2006fully}
Hiroshi Kanayama and Tetsuya Nasukawa.
\newblock Fully automatic lexicon expansion for domain-oriented sentiment
  analysis.
\newblock In {\em Proceedings of the 2006 conference on empirical methods in
  natural language processing}, pages 355--363, 2006.

\bibitem{kraus2017decision}
Mathias Kraus and Stefan Feuerriegel.
\newblock Decision support from financial disclosures with deep neural networks
  and transfer learning.
\newblock {\em Decision Support Systems}, 104:38--48, 2017.

\bibitem{liu2023fingpt}
Xiao-Yang Liu, Guoxuan Wang, and Daochen Zha.
\newblock Fingpt: Democratizing internet-scale data for financial large
  language models.
\newblock {\em arXiv preprint arXiv:2307.10485}, 2023.

\bibitem{maia2018finsslx}
Macedo Maia, Andr{\'e} Freitas, and Siegfried Handschuh.
\newblock Finsslx: A sentiment analysis model for the financial domain using
  text simplification.
\newblock In {\em 2018 IEEE 12th International Conference on Semantic Computing
  (ICSC)}, pages 318--319. IEEE, 2018.

\bibitem{malo2014good}
Pekka Malo, Ankur Sinha, Pekka Korhonen, Jyrki Wallenius, and Pyry Takala.
\newblock Good debt or bad debt: Detecting semantic orientations in economic
  texts.
\newblock {\em Journal of the Association for Information Science and
  Technology}, 65(4):782--796, 2014.

\bibitem{martineau2009delta}
Justin Martineau and Tim Finin.
\newblock Delta tfidf: An improved feature space for sentiment analysis.
\newblock In {\em proceedings of the International AAAI Conference on Web and
  Social Media}, volume~3, pages 258--261, 2009.

\bibitem{mikolov2013distributed}
Tomas Mikolov, Ilya Sutskever, Kai Chen, Greg~S Corrado, and Jeff Dean.
\newblock Distributed representations of words and phrases and their
  compositionality.
\newblock {\em Advances in neural information processing systems}, 26, 2013.

\bibitem{mishev2020evaluation}
Kostadin Mishev, Ana Gjorgjevikj, Irena Vodenska, Lubomir~T Chitkushev, and
  Dimitar Trajanov.
\newblock Evaluation of sentiment analysis in finance: from lexicons to
  transformers.
\newblock {\em IEEE access}, 8:131662--131682, 2020.

\bibitem{pennington2014glove}
Jeffrey Pennington, Richard Socher, and Christopher~D Manning.
\newblock Glove: Global vectors for word representation.
\newblock In {\em Proceedings of the 2014 conference on empirical methods in
  natural language processing (EMNLP)}, pages 1532--1543, 2014.

\bibitem{BloombergReutersDataset2015}
Xiao~Ding Philippe~Remy.
\newblock Financial news dataset from bloomberg and reuters.
\newblock \url{https://github.com/philipperemy/financial-news-dataset}, 2015.

\bibitem{radford2018improving}
Alec Radford, Karthik Narasimhan, Tim Salimans, Ilya Sutskever, et~al.
\newblock Improving language understanding by generative pre-training.
\newblock 2018.

\bibitem{sadia2018overview}
Azeema Sadia, Fariha Khan, and Fatima Bashir.
\newblock An overview of lexicon-based approach for sentiment analysis.
\newblock In {\em 2018 3rd International Electrical Engineering Conference
  (IEEC 2018)}, pages 1--6, 2018.

\bibitem{sarzynska2021detecting}
Justyna Sarzynska-Wawer, Aleksander Wawer, Aleksandra Pawlak, Julia
  Szymanowska, Izabela Stefaniak, Michal Jarkiewicz, and Lukasz Okruszek.
\newblock Detecting formal thought disorder by deep contextualized word
  representations.
\newblock {\em Psychiatry Research}, 304:114135, 2021.

\bibitem{sohangir2018big}
Sahar Sohangir, Dingding Wang, Anna Pomeranets, and Taghi~M Khoshgoftaar.
\newblock Big data: Deep learning for financial sentiment analysis.
\newblock {\em Journal of Big Data}, 5(1):1--25, 2018.

\bibitem{taboada2011lexicon}
Maite Taboada, Julian Brooke, Milan Tofiloski, Kimberly Voll, and Manfred
  Stede.
\newblock Lexicon-based methods for sentiment analysis.
\newblock {\em Computational linguistics}, 37(2):267--307, 2011.

\bibitem{tang2023does}
Ruixiang Tang, Xiaotian Han, Xiaoqian Jiang, and Xia Hu.
\newblock Does synthetic data generation of llms help clinical text mining?
\newblock {\em arXiv preprint arXiv:2303.04360}, 2023.

\bibitem{touvron2023llama}
Hugo Touvron, Thibaut Lavril, Gautier Izacard, Xavier Martinet, Marie-Anne
  Lachaux, Timoth{\'e}e Lacroix, Baptiste Rozi{\`e}re, Naman Goyal, Eric
  Hambro, Faisal Azhar, et~al.
\newblock Llama: Open and efficient foundation language models.
\newblock {\em arXiv preprint arXiv:2302.13971}, 2023.

\bibitem{tripathy2016classification}
Abinash Tripathy, Ankit Agrawal, and Santanu~Kumar Rath.
\newblock Classification of sentiment reviews using n-gram machine learning
  approach.
\newblock {\em Expert Systems with Applications}, 57:117--126, 2016.

\bibitem{vaswani2017attention}
Ashish Vaswani, Noam Shazeer, Niki Parmar, Jakob Uszkoreit, Llion Jones,
  Aidan~N Gomez, {\L}ukasz Kaiser, and Illia Polosukhin.
\newblock Attention is all you need.
\newblock {\em Advances in neural information processing systems}, 30, 2017.

\bibitem{wu2023bloomberggpt}
Shijie Wu, Ozan Irsoy, Steven Lu, Vadim Dabravolski, Mark Dredze, Sebastian
  Gehrmann, Prabhanjan Kambadur, David Rosenberg, and Gideon Mann.
\newblock Bloomberggpt: A large language model for finance.
\newblock {\em arXiv preprint arXiv:2303.17564}, 2023.

\bibitem{zhang2023instruct}
Boyu Zhang, Hongyang Yang, and Xiao-Yang Liu.
\newblock Instruct-fingpt: Financial sentiment analysis by instruction tuning
  of general-purpose large language models.
\newblock {\em arXiv preprint arXiv:2306.12659}, 2023.

\end{thebibliography}

\end{document}